%% file: main.tex
\documentclass[11pt]{article}
\usepackage[final]{acl}   
\usepackage{times}
\usepackage{latexsym}
\usepackage[T1]{fontenc}
\usepackage[utf8]{inputenc}
\usepackage{booktabs}
\usepackage{graphicx}
\usepackage{xcolor}
\usepackage{pgfplots}
\pgfplotsset{compat=1.18}
\usetikzlibrary{positioning,arrows.meta}
\definecolor{outNamed}{HTML}{2A78D6}
\definecolor{outDecline}{HTML}{1BAF7A}
\definecolor{outOther}{HTML}{EB6834}
\definecolor{outNone}{HTML}{B8BCC2}
\definecolor{sysBM}{HTML}{2A78D6}
\definecolor{sysRAG}{HTML}{EB6834}
\definecolor{sysKX}{HTML}{1BAF7A}
\newcommand{\generated}[2]{\IfFileExists{generated/#1}{#2}{\todo{run afribizclaims.report to create generated/#1}}}

\newcommand{\todo}[1]{\textbf{[TODO: #1]}}

\title{Attribute or Abstain: Evaluating Evidence-Grounded Retrieval\\on the Low-Resource African Business Web}

\author{Bakary Gibba \quad Muhammad Abba Mandara \quad Karim Ismail Abd-Ellah Mohamed \\
  Albukhary International University, Malaysia \\
  \texttt{\{bakary.gibba, abba.mandara, karim.mohammed\}@student.aiu.edu.my}}

\begin{document}
\maketitle

\begin{abstract}
Retrieval-augmented systems that research companies are judged by how much they find, yet in markets where public evidence is sparse, a confident answer without support costs more than no answer. We introduce AfriBizClaims, a benchmark of 150 company-research queries across six African markets in English, French and Portuguese, including queries about fictitious companies that a correct system must decline. Each claim a system makes is judged against the page it cites by an AI judge, checked against a second AI judge from another model family and against two human annotators, and abstentions are scored against answerability labels. We compare BM25 over a pooled corpus, standard web retrieval-augmented generation, and a multi-step agent pipeline (Konexium V1), all using the same language model. The two simple baselines are well attributed: 98\% of their judged claims are supported by the page they cite, and they decline every field of every fictitious company. Counting claims about a company other than the one asked about as errors, the agent pipeline's per-query precision is 68\%, and it never answers a question about a named company correctly: it substitutes other firms, including for 5 of the 24 fictitious names, and it is the least stable between runs and the most expensive. In the pipeline we evaluated, more steps between the query and the evidence did not buy more grounded answers. We will release the queries, system outputs, labels and short evidence excerpts.
\end{abstract}

\section{Introduction}

Researching a company in Accra, Nairobi or Maputo often means working with incomplete public evidence. Company registers are only partly online, many company websites are out of date or block automated access, and directories disagree with one another. Retrieval-augmented systems fill such gaps with plausible text: a city taken from a similarly named firm, a website guessed from a name, a sector inferred from a page title. For a business decision, a confident wrong answer is worse than no answer, because it is acted on.

This work grew out of building Konexium, a research platform for African markets whose pipeline attaches a source to every company fact it reports. In early internal tests, few of the sampled claims could be verified independently, which motivated a controlled measurement. We report no figures from those internal tests; every number in this paper comes from the benchmark described here.

Existing attribution benchmarks are built largely on Wikipedia-like, high-resource text \citep{rashkin2023ais,gao2023alce,min2023factscore}. The commercial web of African markets differs: sources are sparse, multilingual and frequently unreachable. We ask:
\textbf{RQ1} How often do standard retrieval pipelines produce company claims that their cited source does not support?
\textbf{RQ2} Do explicit verification gates and abstention reduce unsupported claims, and how much coverage do they cost?
\textbf{RQ3} How does performance vary by country, language and source type?

Our contributions are (1) AfriBizClaims, a benchmark with an annotation protocol for support and answerability, labelled by an AI judge whose labels are checked against human annotators and a second judge; (2) an evaluation that scores attribution and abstention together; and (3) a comparison of three systems, with an error analysis by market, language and source type.

\section{Related Work}

\paragraph{Attribution.} \citet{rashkin2023ais} define attribution to identified sources as a human-judged property of generated text. ALCE \citep{gao2023alce} benchmarks language models that generate text with citations, measuring citation quality alongside correctness, and FActScore \citep{min2023factscore} scores long-form generations by the fraction of their atomic facts supported by a knowledge source. We apply source-level support judgments to company facts, where the cited source is an arbitrary commercial web page rather than an encyclopedia.

\paragraph{RAG evaluation.} The TREC 2024 RAG track and its Ragnar\"ok framework \citep{pradeep2025ragnarok} provide reusable baselines and evaluation for retrieval-augmented generation over a fixed web corpus. We instead evaluate systems that search the live web, and we add unanswerable queries.

\paragraph{Abstention.} Unanswerable questions \citep{rajpurkar2018know} and selective answering under distribution shift \citep{kamath2020selective} show that knowing when not to answer must be measured explicitly. We score abstention per requested field against answerability labels verified on public pages.

\paragraph{African-language retrieval.} AfriCLIRMatrix \citep{ogundepo2022africlirmatrix} and CIRAL \citep{adeyemi2024ciral} provide cross-lingual retrieval test collections for African languages, and AfriBERTa \citep{ogueji2021afriberta} shows that useful multilingual models can be trained on small African-language corpora. Our benchmark is complementary: it targets business facts in English, French and Portuguese, the languages in which much of the African commercial web is written.

\section{Task and Benchmark}

\begin{figure*}[t]
\centering
\begin{tikzpicture}[
  box/.style={draw=gray!70, rounded corners=2pt, align=center, font=\footnotesize, fill=gray!6, inner sep=3pt},
  arr/.style={-{Stealth[length=2mm]}, gray!70, thick}]
\node[box, text width=2.45cm, minimum height=1.6cm] (q) at (0,0) {\textbf{150 queries}\\6 markets; English, French, Portuguese\\84 discovery, 66 verification (24 fictitious)};
\node[box, text width=2.1cm, minimum height=0.75cm] (bm) at (3.4,1.1) {BM25 over a pooled corpus};
\node[box, text width=2.1cm, minimum height=0.75cm] (rag) at (3.4,0) {Web RAG};
\node[box, text width=2.1cm, minimum height=0.75cm, fill=outOther!10] (kx) at (3.4,-1.1) {Multi-step agent pipeline};
\node[box, text width=2.4cm, minimum height=1.6cm] (c) at (6.5,0) {\textbf{Claims}\\a value and one cited URL per field, or no value};
\node[box, text width=3.3cm, minimum height=1.2cm] (j) at (9.95,1.05) {\textbf{Support}\\AI judge on dated page snapshots; checked by 2 humans and a 2nd judge};
\node[box, text width=3.3cm, minimum height=1.2cm] (g) at (9.95,-1.05) {\textbf{Answerability}\\verified agent search plus pooled citations};
\node[box, text width=2.3cm, minimum height=1.6cm] (m) at (13.35,0) {\textbf{Metrics}\\precision, coverage, abstention, yield};
\foreach \s in {bm,rag,kx} \draw[arr] (q.east) -- (\s.west);
\foreach \s in {bm,rag,kx} \draw[arr] (\s.east) -- (c.west);
\draw[arr] (c.east) -- (j.west);
\draw[arr] (c.east) -- (g.west);
\draw[arr] (j.east) -- (m.west);
\draw[arr] (g.east) -- (m.west);
\end{tikzpicture}
\caption{The AfriBizClaims evaluation. All three systems use the same language model; every claim is judged against the page it cites, and empty answers are scored against answerability labels.}
\label{fig:overview}
\end{figure*}
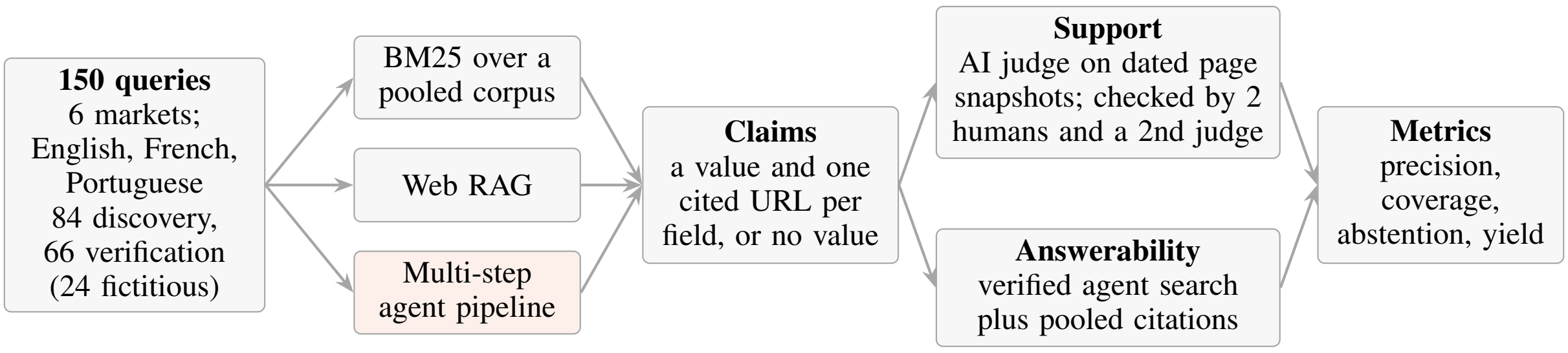

\paragraph{Task.} A system receives a natural-language query and returns, for each company it reports, a value and one cited source URL for each requested field, or no value (an abstention). Fields are company-level only: legal or trading name, country, city, sector, website, registration identifier and operating status. People, e-mail addresses, phone numbers and social profiles are out of scope.

\paragraph{Queries.} AfriBizClaims has 150 queries, 25 for each of Ghana, Kenya, Nigeria and South Africa (English), Senegal (French) and Mozambique (Portuguese). 84 are \emph{discovery} queries (``Find 3 cold-chain logistics companies operating in Kenya''), 24 of which are city-level or niche, where evidence is expected to be thin. 66 are \emph{verification} queries about a named company (``Is Safaricom operating, and what is its website and head-office city?''): 42 name real companies of varying size and 24 name fictitious companies, for which every field is unanswerable. We checked that no fictitious name matched a company in web search. Thirty queries were used as a pilot to fix the protocol; they shaped only the evaluation harness, never a system, and we checked the 120 remaining queries separately: every headline metric there is within 0.04 of the full-set value, so we report the full set.

\paragraph{Annotation.} Each claim is judged against the page it cites, given the question, the value and the URL, without the system name. Support labels for the full set come from an AI judge (OpenAI \texttt{gpt-6.1-sol}, a different model family from the systems' DeepSeek), run with a fixed prompt that restates the human annotation rules on dated text snapshots of each page; two human annotators, working together on one sheet, labelled a random subset of 102 claims (whole pages) plus five quality-check rows with known answers, without system names and without seeing the judge's labels. Their labels agree with the judge on 76 of 102 claims (raw agreement 0.75), but Cohen's $\kappa$ is $-0.04$ (95\% CI $-0.08$--$0.00$), because 18 of the 26 disagreements are verification claims about a company other than the one asked about: the annotators labelled them \textsc{Supported}, since the page does describe that other company, whereas the annotation guide, written before labelling, counts them as \textsc{Wrong-Entity}. Applying the written rule to their labels gives agreement 0.92 and $\kappa=0.78$ (0.36--0.93); because this replaces the annotators' judgement with the guide's rule, we report it only to show where the disagreement lies, not as validation of the judge. The human sheet also missed the checks we had fixed in advance: it quotes no evidence, labels two of the five check rows wrongly (a 404 page and a wrong-company claim as \textsc{Supported}) and uses one label on 95\% of rows. This human round therefore does not validate the judge, and we report it as a failed check. Following the rule we had set for low agreement, we adjudicated a seeded sample of 50 of the judge's 340 \textsc{Wrong-Entity} labels: all 50 concern a company other than the one named in the question or on the page, so no correction was applied. This adjudication applies the same written rule, so it is not independent evidence. The human-labelled subset is reported separately in Appendix~\ref{app:human}. A second AI judge from a third model family (Anthropic \texttt{claude-sonnet-5-5}), given the same snapshots and instructions, agreed with the first on the same 102 claims with $\kappa=0.94$ (95\% CI 0.78--1.00, bootstrap over the 26 pages; raw agreement 0.98); both disagreements were \textsc{Unsupported} against \textsc{Partial}. Both judges labelled all five quality-check rows correctly. Agreement between two models is a robustness check, not a substitute for human validation, since the models may share biases. The judge labelled 1,369 of the 1,405 sampled claims; the other 36 cite pages whose robots.txt disallows fetching. Every \textsc{Supported} or \textsc{Partial} label, human or AI, must quote the sentence that decides it, and quotes are verified against the page automatically. The labels are \textsc{Supported}, \textsc{Partial}, \textsc{Unsupported}, \textsc{Wrong-Entity} (the page is about a different company, or about a company other than the one asked about) and \textsc{Unreachable}, judged on the cited page only, with an excerpt of at most 300 characters. For each verification query and field, answerability was set by a web-search agent and kept only when verified: a field is answerable if a public page states its value and that page, fetched independently, contains the quoted statement; it is unanswerable if no such page was found. As in pooled evaluation, a page that a system cited also counts as evidence when the judge labelled the claim \textsc{Supported} and its quote was verified on the page. Of the 252 fields about real companies, 198 are answerable (143 from the agent search, 55 from pooled citations), 32 are unanswerable and 22 could not be settled and are left out. Every field about a fictitious company (144) is unanswerable by construction.

\section{Systems}

All three systems use the same language model, DeepSeek-V4.1-Flash (\texttt{deepseek-flash}) with reasoning disabled, so differences come from retrieval and verification.

\paragraph{BM25.} Pyserini \citep{lin2021pyserini} BM25 \citep{robertson2009bm25} over a pooled corpus: the pages fetched by the RAG baseline plus every page cited by the Konexium run. The top five passages go to the model with a fixed prompt that requires one cited source per value and a null value when no source states a field.

\paragraph{Web RAG.} The query is sent to a web search API (Serper, Google results localised by country and language); the top five results are fetched and their main text extracted, and the same prompt and model produce the answer. There is no verification step.

\paragraph{Konexium V1.} A multi-step agent pipeline, frozen at one commit, from a pre-launch product. It plans a search from the request, searches several engines and crawls candidate websites, and is designed to assess entity identity and country, apply quality gates, store each extracted fact with its source, and abstain when a gate fails. It has no city field; we export the first entry of its free-text location list as the city. Its assistant receives the query text unchanged. When it asks a clarifying question instead of acting, a fixed reply restates facts already in the query; we report how often this was needed. Konexium's built-in fallbacks to other model vendors were routed to the same model, and it started from an empty database, so no evidence collected earlier was reused.

\paragraph{Fetching.} The baselines respect robots.txt and identify themselves with a contact URL; blocked pages are simply missing sources. All runs were made on 3 November 2026: an evaluation round, which is annotated, and a second identical round used only to measure stability.

\section{Metrics}

\textbf{Citation precision} is the share of judged claims labelled \textsc{Supported} (strict) or \textsc{Supported} or \textsc{Partial} (lenient); \textsc{Unreachable} claims are excluded and reported separately as the \textbf{unreachable rate}, so a system is neither rewarded nor punished for sites that block automated access. The \textbf{unsupported-claim rate} counts \textsc{Unsupported} and \textsc{Wrong-Entity} claims among judged claims. On verification queries, \textbf{coverage} is the share of answerable (query, field) pairs answered with a supported claim, and \textbf{abstention precision and recall} treat an empty answer as a prediction that the field is unanswerable. For discovery queries, \textbf{discovery yield} is the mean share of the requested companies returned with a supported sector claim. Because only the first company each system returned per discovery query was annotated, yield is estimated per query as the share of the request returned times whether the annotated company's sector claim is supported; verification claims were annotated in full. Latency per query and cost (model tokens, search requests) are reported separately and never combined with quality. Because one answer can contain many claims, we report strict precision both per claim and averaged per query (Prec.$_Q$); the per-query figure is the headline, so that a few long answers cannot dominate it. Intervals are 95\% bootstrap intervals over queries.

\paragraph{Protocol changes.} Four parts of the protocol changed during the study, all recorded with dates in a decision log that will be released with the data. Annotation was first planned for every claim by humans; to keep it tractable we annotated only the first company per discovery answer, and labels then came from the AI judge validated as above. Answerability, first planned as human labels, came from the verified agent search, to which we added pooled evidence after seeing that the verifier left a quarter of the fields unsettled. The per-query headline was chosen after the labels were seen; both figures are shown throughout.

\section{Results}

\begin{table*}[t]
\centering\small
\generated{table1.tex}{\input{generated/table1.tex}}
\caption{Main results per system on the evaluation round. Claims: claims made, including unannotated ones; Prec.$_Q$: strict citation precision averaged per query; Prec.: strict citation precision per claim; Prec.$_L$: lenient; Unsup.: unsupported-claim rate; Unreach.: share of claims whose source could not be opened; Cov.: coverage of answerable verification fields; Abst.P/R: abstention precision and recall; Yield: discovery yield.}
\label{tab:main}
\end{table*}

\begin{table*}[t]
\centering\small
\generated{table2-language.tex}{\input{generated/table2-language.tex}}
\caption{Results by query language. Breakdowns by country and source type will be included in the data release.}
\label{tab:language}
\end{table*}

\begin{figure}[t]
\centering
\generated{fig1.csv}{%
\begin{tikzpicture}
\begin{axis}[width=0.97\columnwidth, height=0.72\columnwidth,
  xlabel={Coverage (answerable fields answered with support)},
  ylabel={Unsupported-claim rate},
  xmin=0, xmax=1, ymin=0, ymax=1,
  grid=major, grid style={gray!20}, axis line style={gray!60}, tick style={gray!60},
  label style={font=\small}, tick label style={font=\footnotesize}]
\addplot[only marks, mark=*, mark size=3pt, color=sysBM,
  nodes near coords={BM25}, every node near coord/.style={font=\footnotesize, text=black, anchor=south east, yshift=2pt}]
  table[x=coverage, y=unsupported_rate, col sep=comma, restrict expr to domain={\coordindex}{0:0}]{generated/fig1.csv};
\addplot[only marks, mark=square*, mark size=3pt, color=sysKX,
  nodes near coords={Konexium}, every node near coord/.style={font=\footnotesize, text=black, anchor=west, xshift=2pt}]
  table[x=coverage, y=unsupported_rate, col sep=comma, restrict expr to domain={\coordindex}{1:1}]{generated/fig1.csv};
\addplot[only marks, mark=triangle*, mark size=3.5pt, color=sysRAG,
  nodes near coords={Web RAG}, every node near coord/.style={font=\footnotesize, text=black, anchor=south west, yshift=2pt}]
  table[x=coverage, y=unsupported_rate, col sep=comma, restrict expr to domain={\coordindex}{2:2}]{generated/fig1.csv};
\end{axis}
\end{tikzpicture}}
\caption{Coverage against unsupported-claim rate, one point per system. Lower right is better.}
\label{fig:tradeoff}
\end{figure}
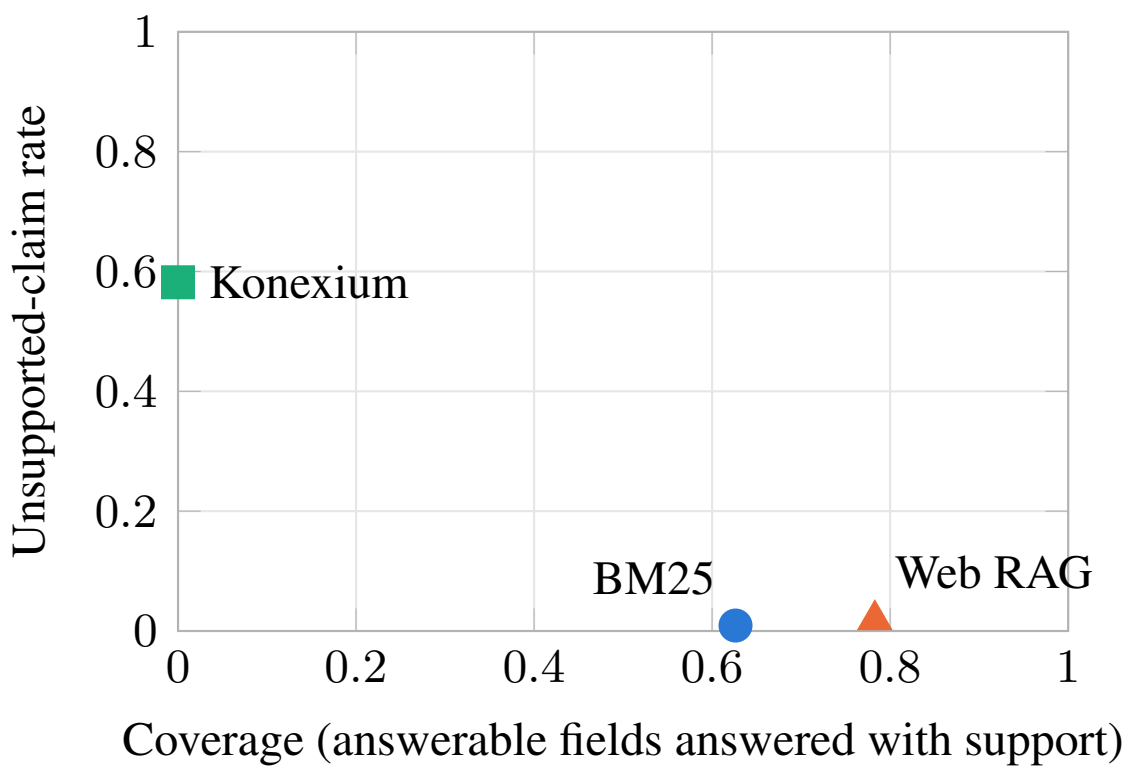

\paragraph{Attribution (RQ1).} The two simple systems are well attributed (Table~\ref{tab:main}). Averaged per query, 98\% of the judged claims of BM25 (95\% CI 0.95--0.99) and of web RAG (0.96--1.00) are supported by the page they cite, and fewer than 2\% are unsupported. The agent pipeline reaches 0.68 (0.60--0.76) per query and 0.39 per claim. The two figures differ because its failures are concentrated: on verification queries none of its claims is supported, while on discovery queries its per-query precision is 0.84, against 0.97 for BM25 and 0.98 for web RAG. Much of that discovery gap lies in the city field, which we take from the pipeline's free-text location list: without city claims, its discovery precision is 0.91, against 0.98 and 0.99. The baselines' near-ceiling precision is partly by construction, since they answer from the five passages or pages they retrieve under a prompt that requires a source for every value; their remaining errors concern which entity or place a page describes, not whether a source was cited.

\begin{figure}[t]
\centering
\begin{tikzpicture}
\begin{axis}[xbar stacked, width=0.85\columnwidth, height=0.5\columnwidth,
  xmin=0, xmax=66, bar width=11pt, y=0.62cm,
  symbolic y coords={Konexium,Web RAG,BM25}, ytick=data, enlarge y limits=0.3,
  xlabel={Verification queries (of 66)}, xtick={0,11,22,33,44,55,66},
  axis line style={gray!60}, tick style={gray!60}, label style={font=\small}, tick label style={font=\footnotesize},
  legend style={font=\scriptsize, at={(0.5,1.03)}, anchor=south, legend columns=2, draw=none, /tikz/every even column/.append style={column sep=6pt}},
  legend cell align=left, every axis plot/.append style={draw=white, line width=1pt}]
\addplot[fill=outNamed] coordinates {(34,BM25) (40,Web RAG) (0,Konexium)};
\addplot[fill=outDecline] coordinates {(24,BM25) (24,Web RAG) (19,Konexium)};
\addplot[fill=outOther] coordinates {(0,BM25) (0,Web RAG) (20,Konexium)};
\addplot[fill=outNone] coordinates {(8,BM25) (2,Web RAG) (27,Konexium)};
\legend{About the named company, Declined a fictitious company, About another company, No answer for a real company}
\end{axis}
\end{tikzpicture}
\caption{What each system did with the 66 verification queries. The agent pipeline never answered about the company named in the query.}
\label{fig:verification}
\end{figure}
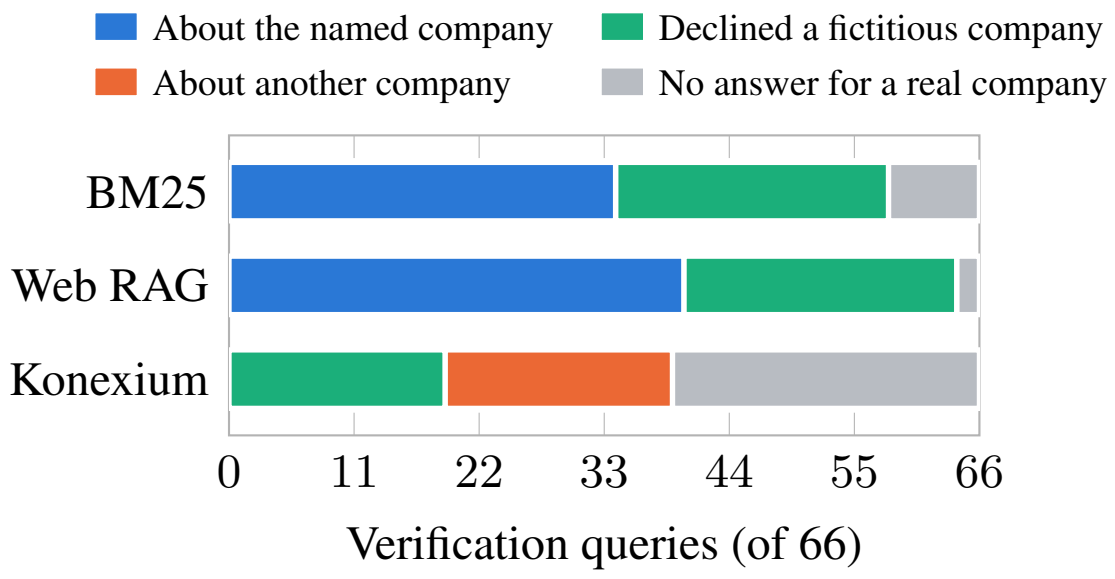

\paragraph{Coverage and abstention (RQ2).} Web RAG answers 78\% of the answerable verification fields with a supported value (0.71--0.85) and BM25 63\% (0.52--0.73). The agent pipeline answers none (Figure~\ref{fig:verification}). On 45 of the 66 verification queries it did not start any research; on the other 21 it ran a generic search, and on 20 of these it reported other companies, sometimes more than a dozen, every one judged \textsc{Wrong-Entity}. Its abstention recall in Table~\ref{tab:main} therefore mostly counts requests it never acted on: on the 21 verification queries where it ran, its abstention recall is 0.55 and its abstention precision 0.57. On the 24 fictitious companies the baselines never stated a value, while the agent pipeline reported other, real companies for 5. These five are the only fictitious names it researched at all; the other 19 are among the queries on which it started no research, so it never declined a fictitious company after searching for it. Its abstentions are also less informative: when it gives no value, the field is in fact unanswerable 52\% of the time, against 73\% for BM25 and 85\% for web RAG, so its silence mostly reflects the requests it could not act on rather than missing evidence. The verification gates in this pipeline did not lower the unsupported-claim rate (Figure~\ref{fig:tradeoff}): they assess each candidate company, but in our runs nothing tied the candidates to the company the user named. On discovery the three systems are close: the estimated share of requested companies returned with a supported sector is 0.67 for web RAG and the agent pipeline and 0.62 for BM25.

\paragraph{Language and market (RQ3).} Both baselines keep a per-query precision of at least 0.94 in every language and market (Table~\ref{tab:language}). For the agent pipeline, per-query precision is 0.77 on French queries, 0.70 on Portuguese and 0.66 on English; the English figure includes most of the verification queries on which it substituted other firms. Coverage varies more than precision: BM25 covers 31\% of answerable Ghanaian fields and 78\% of Kenyan ones, which reflects what its pooled corpus contains rather than the market itself. With 25 queries per market most intervals overlap, so we do not rank markets.

\paragraph{Error analysis.} We drew 30 failed claims at random (9 from BM25, 10 from web RAG, 11 from the agent pipeline) and assigned each a cause. Ten of the 11 agent-pipeline failures are verification answers about another firm, and one is a discovery answer that gives the country as the city. The substituted firms are almost all consulting, outsourcing or company-registration services (for the fictitious ``Benue Crestline Agro Industries Ltd'', 17 such firms), which suggests that the pipeline read a question such as ``Is X registered and operating in Nigeria?'' as a request for that sector. The 19 baseline failures fall into four groups. Six take a location that the page gives for something else, as the city: a factory, a shop, a job posting or a place in the company's name. Five are name collisions on directory pages (a ``Dana Pharmaceuticals'' answer citing a list that names Dana Drugs Limited). Five cite the same page, which returned HTTP 404 when judged. Three report a value the page does not state, such as an operating status read from a site's maintenance notice. The same pages recur across the two baselines, because the BM25 corpus contains the pages that web RAG retrieved. A separate look at the agent pipeline's 66 discovery claims that were not supported shows a format effect: its city field usually holds a full street address (61 of its 72 judged city claims) or only the country (8), and 22 of the 66 failures are addresses not fully stated on the page or a country given as the city. Another 19 are pages we could not open when judging (11 connection errors, 8 refusals with HTTP 403 or 429), which count toward the unreachable rate but not against precision.

\paragraph{Stability, cost and latency.} Running each system twice on the same queries shows how reproducible its answers are, independently of correctness (Table~\ref{tab:stability}). BM25 over a fixed index and web RAG return largely the same companies in both runs; Konexium's two runs share only about a third of their discovery companies and make the same answer-or-abstain decision on 68\% of verification fields. Konexium is also the slowest and most expensive in model calls, because it researches many candidates per query.

\paragraph{Clarifying questions.} Konexium's assistant asked before acting on 99 of the 150 queries in the evaluation round. The pattern depends on language: it needed the fixed reply for 7 of 56 English discovery queries but for all 28 French and Portuguese ones, where it did not recognise the country named in the query. It has no action for verifying one named company, so on 45 of the 66 verification queries it ended without starting any research; its replies there declined to state facts from memory, which our metrics count as abstentions.

\begin{table}[t]
\centering\small
\begin{tabular}{lrrrr}
\toprule
System & Jacc. & Same & Sec. & Cost \\
\midrule
BM25 & 0.95 & 0.99 & 1.1 & $\approx$0.05 \\
Web RAG & 0.80 & 0.97 & 20 & 0.09 \\
Konexium & 0.36 & 0.68 & 44 & $\approx$0.93 \\
\bottomrule
\end{tabular}
\caption{Run-to-run stability and cost over the 150 queries. Jacc.: mean Jaccard overlap of the companies returned per discovery query between two runs; Same: share of verification fields answered in both runs or abstained in both; Sec.: median seconds per query; Cost: model cost in US dollars per 150 queries. Web RAG also made 150 search requests.}
\label{tab:stability}
\end{table}

\section{Conclusion}

AfriBizClaims measures whether company facts about six African markets are supported by the pages that systems cite, and whether systems decline when no public evidence exists. On this benchmark, single-step retrieval with a strict citation prompt was well attributed and declined every fictitious company, while the multi-step agent pipeline we evaluated, which has verification gates, was less precise, never answered a named-company question correctly, and was less stable and more costly. Most of its failures came less from weak evidence than from losing track of the request: questions about one company became generic searches, and the gates then checked the wrong candidates. These results come from one agent pipeline and one model, so for agent pipelines in general they are hypotheses rather than conclusions. We take two lessons for evaluating such systems. First, entity fidelity, whether an answer concerns the entity that was asked about, should be measured alongside citation support, because a page can support a claim perfectly and still be about the wrong company. Second, abstention needs answerability labels: an empty answer that comes from a request the system could not handle looks like caution but is not. The 150 queries, outputs, labels and excerpts will be released so that other systems, including later versions of the one evaluated here, can be measured on the same footing.

\section*{Limitations}

The benchmark is small: 150 queries in six markets and three languages. It is a snapshot of a changing web, so pages cited today may change or disappear; we will release dated excerpts for this reason. The BM25 corpus is pooled from the other systems' retrievals, which bounds what BM25 can find and makes it partly an oracle-retrieval baseline. Answerability labels also draw on pages the systems cited, so a fact that no system and no agent search found counts as unanswerable; 55 of the 198 answerable fields rest on pooled citations, and 22 fields were left unsettled. The error analysis covers 30 sampled failures assigned to causes by the first author, not by an independent annotator. Support labels come from an AI judge. Its agreement with the human annotators rests on 102 claims from 26 pages, the human sheet did not meet our quality checks, and agreement is high only once the written wrong-entity rule is applied, so the judge's labels are not validated by an independent human check, only by agreement with a second AI judge; the judge may also err systematically on page types that subset does not cover. Konexium's citation precision depends on that rule: on the human-labelled claims, reading support from the page alone would raise it from 0.50 to 0.97, while its coverage and abstention, which count only answers about the named company, would not change. Pages whose robots.txt disallows fetching were not judged. Konexium was evaluated through its conversational assistant with a fixed reply protocol; a user who answers questions differently may get different results. One model was used throughout, and results may differ with others.

\section*{Ethics Statement}

The first author founded Konexium, one of the evaluated systems. To allow the comparison to be checked, we will release every system's outputs, the labels and the evaluation code, and the Konexium version is fixed by commit; the benchmark was never used to tune it. The two co-authors who labelled the validation subset are not part of Konexium; they labelled a blinded sheet without system names and without seeing the judge's labels, and a second AI judge from another model family was checked against both. The benchmark covers company-level facts only. Konexium's pipeline, as frozen, also extracts names of company decision-makers; these stayed in a local evaluation database that no export read, are not part of any release, and will be deleted when the study ends. Pages were fetched only where robots.txt allows; no login, CAPTCHA or other access control was bypassed. Only excerpts of at most 300 characters will be released, never full pages.

\section*{Acknowledgements}

AI assistants were used for code and drafting; the authors reviewed all content and are responsible for it.

\bibliography{references}
\bibliographystyle{acl_natbib}

\appendix
\section{Human-labelled subset}
\label{app:human}

Table~\ref{tab:human} gives strict citation precision per claim on the 102 claims labelled by the human annotators, with their labels as given, with the written wrong-entity rule applied to them, and with the AI judge's labels on the same claims. The two readings differ only for the agent pipeline, whose verification answers concern companies other than the one asked about.

\begin{table}[h]
\centering\small
\begin{tabular}{lrrrr}
\toprule
System & Claims & Human & Human, rule & Judge \\
\midrule
BM25 & 30 & 0.97 & 0.97 & 0.94 \\
Web RAG & 32 & 0.97 & 0.97 & 1.00 \\
Konexium & 38 & 0.97 & 0.50 & 0.50 \\
\bottomrule
\end{tabular}
\caption{Strict precision per claim on the human-labelled subset. Claims: judged claims (unreachable excluded) under the human labels; the judge judged 31 and 33 for BM25 and web RAG.}
\label{tab:human}
\end{table}

\end{document}

%% file: generated/table1.tex
\begin{tabular}{lrrrrrrrrrr}
\toprule
System & Claims & Prec.$_Q$ & Prec. & Prec.$_L$ & Unsup. & Unreach. & Cov. & Abst.P & Abst.R & Yield \\
\midrule
BM25 & 623 & 0.98 & 0.98 & 0.99 & 0.01 & 0.00 & 0.63 & 0.73 & 1.00 & 0.62 \\
Web RAG & 692 & 0.98 & 0.98 & 0.98 & 0.02 & 0.00 & 0.78 & 0.85 & 0.99 & 0.67 \\
Konexium & 1190 & 0.68 & 0.39 & 0.42 & 0.58 & 0.02 & 0.00 & 0.52 & 0.89 & 0.67 \\
\bottomrule
\end{tabular}

%% file: generated/table2-language.tex
\begin{tabular}{lrrrrrrrrrr}
\toprule
System & Claims & Prec.$_Q$ & Prec. & Prec.$_L$ & Unsup. & Unreach. & Cov. & Abst.P & Abst.R & Yield \\
\midrule
BM25 / en & 458 & 0.98 & 0.99 & 0.99 & 0.01 & 0.00 & 0.65 & 0.75 & 1.00 & 0.70 \\
BM25 / fr & 113 & 0.94 & 0.93 & 0.98 & 0.02 & 0.00 & 0.62 & 0.71 & 1.00 & 0.67 \\
BM25 / pt & 52 & 1.00 & 1.00 & 1.00 & 0.00 & 0.04 & 0.53 & 0.69 & 1.00 & 0.26 \\
Web RAG / en & 472 & 0.98 & 0.98 & 0.98 & 0.02 & 0.00 & 0.76 & 0.83 & 0.99 & 0.68 \\
Web RAG / fr & 122 & 0.97 & 0.97 & 0.99 & 0.01 & 0.00 & 0.88 & 0.91 & 1.00 & 0.69 \\
Web RAG / pt & 98 & 0.98 & 0.98 & 0.98 & 0.02 & 0.03 & 0.80 & 0.88 & 1.00 & 0.62 \\
Konexium / en & 909 & 0.66 & 0.34 & 0.37 & 0.63 & 0.02 & 0.00 & 0.52 & 0.86 & 0.70 \\
Konexium / fr & 134 & 0.77 & 0.77 & 0.79 & 0.21 & 0.03 & 0.00 & 0.52 & 1.00 & 0.67 \\
Konexium / pt & 147 & 0.70 & 0.49 & 0.49 & 0.51 & 0.03 & 0.00 & 0.52 & 0.86 & 0.60 \\
\bottomrule
\end{tabular}